\documentclass[fleqn,10pt]{wlscirep}
\usepackage[utf8]{inputenc}
\usepackage[T1]{fontenc}
\usepackage{lineno}
\usepackage{float}
\usepackage{tabularx}
\usepackage{booktabs}
\usepackage{subcaption}

\title{User eXperience Perception Insights Dataset (UXPID): Synthetic User Feedback from Public Industrial Forums}

\author[1, $\dag$]{Mikhail Kulyabin}
\author[1, $\dag$]{Jan Joosten}
\author[1,2]{Choro Ulan uulu}
\author[1]{Nuno Miguel Martins Pacheco}
\author[1]{Fabian Ries}
\author[1]{Filippos Petridis}
\author[2,3,*]{Jan Bosch}
\author[4]{Helena Holmström Olsson}

\affil[1]{Siemens AG, Erlangen, Germany}
\affil[2]{Department of Mathematics and Computer Science, Eindhoven University of Technology, Eindhoven, Netherlands}
\affil[3]{Department of Computer Science and Engineering, Chalmers University of Technology, Gothenburg, Sweden}
\affil[4]{Department of Computer Science and Media Technology, Malmö University, Malmö, Sweden}
\affil[$\dag$]{These authors contributed equally to this work}
\affil[*]{corresponding author: Jan Bosch (jan.bosch@chalmers.se)}

\begin{abstract}
Customer feedback in industrial forums offers rich but underexplored insights into real-world product experience. Yet systematic analysis remains challenging due to unstructured, domain-specific content and the scarcity of high-quality labeled datasets. This paper presents the User eXperience Perception Insights Dataset (UXPID), a collection of 7130 synthesized and anonymized user feedback branches extracted from a public industrial automation forum. Each JSON record contains multi-post comments enriched with metadata and annotated by a large language model (LLM) for UX insights, user expectations, severity ratings, sentiment, and topic classifications. UXPID is designed to facilitate research in user requirements, user experience (UX) analysis, and AI-driven feedback processing, particularly where privacy and licensing restrictions limit access to real-world data. It supports the training and evaluation of transformer-based models for tasks such as issue detection, sentiment analysis, and requirements extraction in technical forums, providing a valuable resource for advancing NLP methods within industrial product support and software engineering domains.
\end{abstract}
\begin{document}

\flushbottom
\maketitle
        
\thispagestyle{empty}

\section*{Background \& Summary}

User feedback in technical forums provides valuable insights into real-world experiences, usability issues, and feature requests for software and hardware products \cite{maalej2025automated}. Researchers are increasingly using natural language processing (NLP) and transformer-based models to analyze such feedback at scale \cite{zhang2025less}. However, progress remains constrained by the limited availability of datasets that capture structured insights from real-world user discussions.
Existing datasets for requirements engineering and user feedback analysis often lack contextual discussion structures, severity annotations, or semantic insights derived from user conversations (see Table 1). These limitations restrict the development and evaluation of models that aim to extract actionable requirements and UX insights from forum-style discussions.

Human-centered design (HCD), as defined by the ISO 9241-210:2019 standard \cite{ISO9241-210}, is an iterative approach that prioritizes users’ needs, contexts, and experiences in system development. Unlike traditional requirements engineering (RE), which focuses on eliciting and documenting functional and non-functional requirements \cite{lopez_2023_re}, HCD emphasizes continuous user involvement throughout the development lifecycle. This perspective highlights the importance of systematically analyzing user feedback to inform product design and engineering decisions \cite{sommerville1997requirements, Harte2017}.

Software and product development involve the design, implementation, and delivery of systems that address user and business requirements. Within this context, user experience (UX) design focuses on improving the quality of user interactions by emphasizing usability, accessibility, and satisfaction \cite{Sauer02102020}.

Within these domains, user feedback plays a central role in shaping requirements, guiding design decisions, and evaluating product performance and usability \cite{maalej2025automated, soares_user_2011}. Such feedback appears in support tickets, product forums, online reviews, surveys, and social media discussions. These sources provide valuable insights into user expectations and challenges. However, the data is typically unstructured, noisy, and distributed across multiple platforms, which makes systematic analysis difficult.

Before the development of natural language processing (NLP), product managers had to manually analyze user comments to identify issues such as product bugs, negative reviews, and requests for new features. Manual analysis is time-consuming and difficult to scale.

In recent years, artificial intelligence (AI), particularly NLP, has advanced rapidly. NLP enables machines to understand, interpret, and generate human language by combining techniques from computational linguistics, statistics, machine learning, and deep learning \cite{kang2020natural, hirschberg2015advances}. These advances allow systems to process large volumes of user feedback and extract actionable insights \cite{just2024natural, maalej2025automated}.

Although recent advances in NLP and transformer-based models have shown promise in automating feedback analysis, progress is limited by the scarcity of high-quality task-specific training data\cite{laurer2024less}. While NLP encompasses a broad set of techniques for analyzing and understanding human language, transformer models represent a specific deep learning architecture that has dramatically improved the performance of many NLP tasks. For this, transformer models require large and well-labeled datasets to achieve strong performance in domain-specific tasks \cite{zhang2025less}. However, publicly available datasets that reflect the insights of real-world user discussions, such as those found in product forums, are scarce. This limits the ability of researchers and practitioners to train or fine-tune models for tasks such as issue detection, feature request classification, sentiment analysis, and requirements extraction for users on platforms like technical forums.

To address this gap, we present the User eXperience Perception Insights Dataset (UXPID), a synthetic dataset that reflects forum-style user discussion summaries in a software and hardware product context. The dataset is designed to support the training and fine-tuning of transformer-based models for downstream tasks relevant to RE and UX, including feedback classification, topic clustering, voice-of-user clustering, and requirement identification. The data set includes key properties such as labeled examples, conversational branch summaries, and UX insights, and metadata, and is constructed to reflect the linguistic and structural characteristics of real user forums while avoiding privacy and licensing concerns.

In this work, we collected a comprehensive dataset from an international technical forum, enriched it through semantic analysis and automated insight generation with large language models (LLMs), validated the annotations through human assessment, and implemented robust anonymization techniques to protect user privacy. The resulting dataset is research-ready and designed to support diverse applications in requirements engineering and natural language processing.

The main contributions of this work include:
\begin{enumerate}[label=\textbf{(\arabic*)}, leftmargin=*, itemsep=2pt, topsep=2pt]
  \item A novel dataset (\textsc{UXPID}) that captures user comments and associated metadata from an international technical forum, extending the scope of existing datasets in requirements engineering.
  \item Rich semantic structure, including automated insights, topic clustering, and severity annotations, which provide contextual depth beyond what prior resources offer.

  \item Benchmark baselines for classification tasks using transformer-based models, offering a solid reference point for future research and applied work.
\end{enumerate}

The remainder of this paper is organized as follows. First, we review existing open-access datasets in requirements engineering and discuss their limitations, thereby motivating the need for the proposed UXPID dataset. Next, we introduce the construction and properties of the UXPID dataset in detail. Finally, we present a technical validation of the dataset through classification experiments using a transformer-based model.

\subsection*{Open-access datasets}

In this section we describe several open-access datasets commonly used for research on technical forums, requirements engineering, and user feedback. These datasets differ in domain, annotation type, and data structure, providing resources for tasks such as relevance classification, requirements extraction, and sentiment analysis. Their diversity supports comparative studies and demonstrates the need for more specialized datasets in industrial contexts.

\setlength{\tabcolsep}{4pt}
\renewcommand{\arraystretch}{1.2}

\begin{table}[h!]
\small
\centering
\caption{Comparative analysis of open-access datasets related to this work.}
\begin{tabularx}{\textwidth}{@{}l>{\raggedright\arraybackslash}p{2.3cm}>{\raggedright\arraybackslash}p{2.5cm}>{\raggedright\arraybackslash}X>{\raggedright\arraybackslash}X@{}}
\toprule
\textbf{Dataset Name} & \textbf{Domain / Source} & \textbf{Size / Stats} & \textbf{Content Description} & \textbf{Purpose / Use Case} \\
\midrule
\textbf{TFThs} \cite{osman_quality_2019} & Forum Threads & Large collection (exact size not specified) & Post–reply pairs with class labels indicating relevance to the initial post. & Relevance classification, thread analysis \\
\midrule
\textbf{TechQA} \cite{castelli2019techqa} & IBM Technical Support & 600 items: 310 development set, 490 evaluation pairs & Question–answer pairs related to IBM technical support. & QA system evaluation, domain-specific QA \\
\midrule
\textbf{FR-NFR} \cite{sonali_fr_nfr_dataset_2024} & Requirements Engineering & 6118 requirements: 3964 functional, 2154 non-functional & Software development requirements labeled as functional or non-functional. & RE classification, functional vs. non-functional detection \\
\midrule
\textbf{PURE} \cite{ferrari2017pure} & Natural Language Requirements & 34268 sentences from 79 public documents & English sentences from natural language requirements documents. & NLP tasks, requirements engineering \\
\midrule
\textbf{Bozyigit et al.} \cite{Bozyigit_2023_r9j6-nd62-23} & Functional Requirements & 120 text files & Functional requirements in English across domains like software, business, education. & Model generation from text requirements \\
\midrule
\textbf{PROMISE} \cite{mekala2021classifying} & Software Projects & 101 software projects & Repository of software defect data across multiple projects. & Defect classification, software engineering research \\
\midrule
\textbf{DOSSPRE} \cite{kadebu2023classification} & Requirements Classification & 1317 requirements & Software requirements across categories including security, functional, and non-functional. & Requirement classification \\
\midrule
\textbf{NEODATASET} \cite{neo_user_2024} & Agile Software Development & 34 projects, 40014 user stories, 163897 story points & User stories with assigned story points from real-world software projects. & Agile estimation, productivity analysis \\
\midrule
\textbf{UXPID} \cite{Kulyabin2025UXPID} & Technical Forum & 7130 branches with 32k anonymized user comments & Customer feedback on the automatization products & Training of NLP models to solve classification problems \\
\bottomrule
\end{tabularx}
\label{tab:datasets}
\end{table}

The TFThs dataset \cite{osman_quality_2019} contains a large quantity of posts-replies pairs. These forum threads have class labels that reflect relevance to the initial post. The TechQA dataset \cite{castelli2019techqa} is a question-answer dataset for an IBM technical support domain. It contains 600 items, 310 development set, and 490 evaluation question-answer pairs. The FR-NFR dataset \cite{sonali_fr_nfr_dataset_2024} includes data of requirement engineering in software development. The dataset comprises 6118 requirements, of which 3964 are functional and 2154 are non-functional. The PURE dataset \cite{ferrari2017pure} comprises 34268 English sentences that can be utilized for NLP tasks, such as requirements engineering. It consists of 79 publicly available natural language requirements documents online. The Bozyigit et al. dataset for text requirements to models \cite{Bozyigit_2023_r9j6-nd62-23} consists of 120 text files that describe functional requirements written in English. The settings differ across contexts, including software systems, education, business, and hospitality. The PROMISE repository \cite{mekala2021classifying} contains defect data of 101 software projects. The projects have different use cases. The DOSSPRE dataset \cite{kadebu2023classification} includes 1317 software requirements. The requirements range from security requirements to functional and other non-functional requirements. The NEODATASET \cite{neo_user_2024} contains 34 software development projects with 40014 user stories. The total story points is 163897. Table~\ref{tab:datasets} summarizes their key properties.

Despite the availability of several datasets relevant to requirements engineering and software development (e.g., TFThs \cite{osman_quality_2019}, TechQA \cite{castelli2019techqa}, FR-NFR-dataset \cite{sonali_fr_nfr_dataset_2024}, PURE \cite{ferrari2017pure}, Bozyigit et al. \cite{Bozyigit_2023_r9j6-nd62-23}, PROMISE \cite{mekala2021classifying}, DOSSPRE \cite{kadebu2023classification}, and NEODATASET \cite{neo_user_2024}), each resource exhibits critical limitations that restrict their applicability. Some datasets are too small to support robust model training (e.g., TechQA, Bozyigit), while others, though larger in scale (e.g., PURE, NEODATASET), lack depth into specific branches of requirements or do not incorporate contextual framing beyond raw requirements. Existing datasets also typically omit severity or criticality annotations, preventing the development of risk-sensitive models. Moreover, they provide little to no support for scenario-based reasoning or structured insight generation, which are essential for practical applicability. Finally, current datasets rarely implement explicit topic clustering, limiting their use for comparative analyses or advanced NLP techniques such as transformer-based topic modeling.

\section*{Methods}

The UXPID dataset is composed of user comments - including both feedback and queries - carefully extracted from discussion branches hosted on a company-operated forum dedicated to automation product hardware and software components. While the forum is publicly accessible, we do not disclose its name or provide direct URLs to reduce the risk of re-identification and traceability of individual contributors and posts, even after anonymization and synthesis. Use of the forum content is governed by the forum’s terms of Use/Forum Terms, which participants must accept; these terms state that authors retain copyright in their posts while granting the owner a broad, perpetual, and irrevocable license to use, reproduce, modify, distribute, and display submitted content across media. Each conversational branch was systematically analyzed using LLM, enabling the extraction of summarized insights. These insights include detailed problem descriptions, user expectations, the perceived severity of reported issues, and overall sentiment expressed within the branches. By capturing both the technical and experiential aspects of user interactions, UXPID provides a view of product usage and customer experience in industrial automation contexts.

\begin{figure}[H]
    \centering
    \includegraphics[width=0.7\columnwidth]{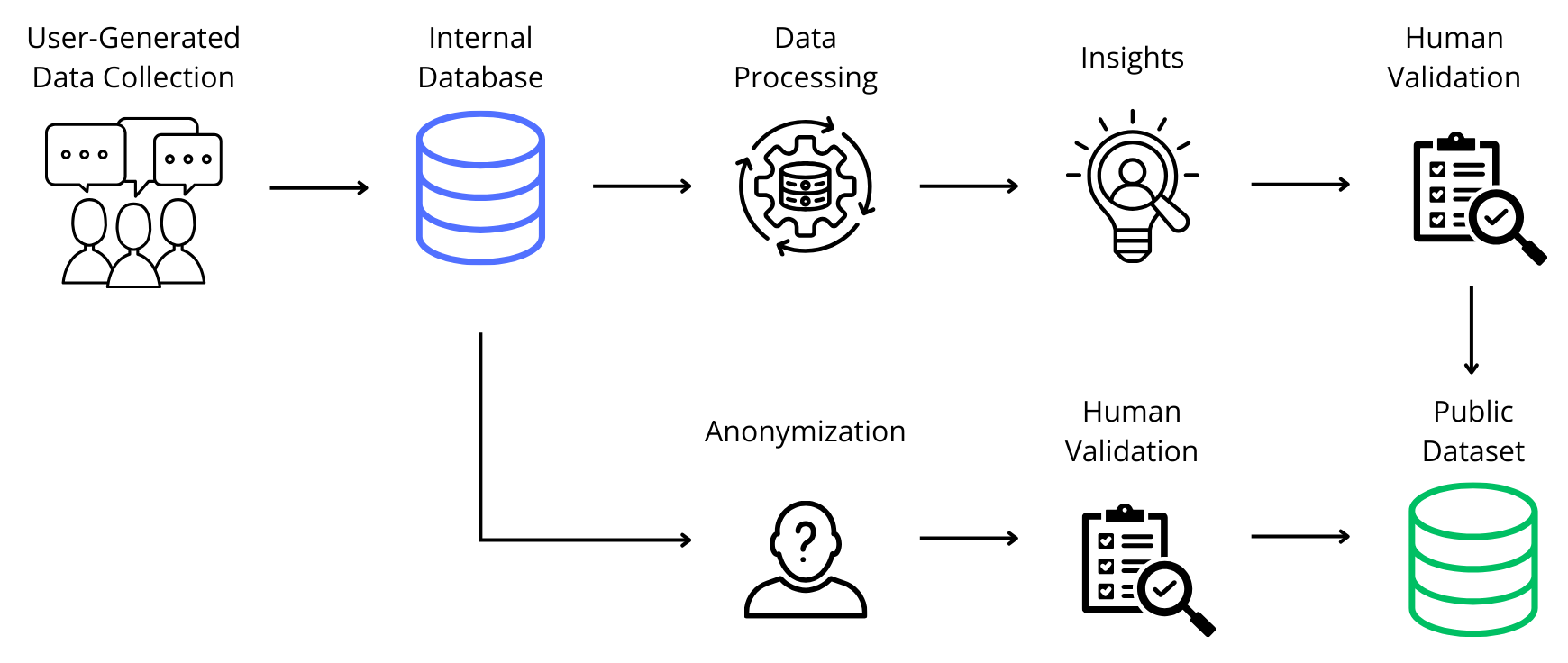}
    \caption{Overview of dataset creation process}
    \label{fig:pipeline}
\end{figure}

Fig.~\ref{fig:pipeline} illustrates the general process for dataset creation. User comments were collected from the company open public technical forum. Endpoints enable the structured storage of metadata, including user id, date, and time of posting, titles, and the content of the comments themselves. For internal processing and analysis, the data are stored in JavaScript object notation (JSON) format.

It is important to emphasize the sequential nature of the pipeline to avoid any misconception about circularity between annotation and synthesis. In the first step, all semantic annotations were extracted by the LLM directly from the original, unmodified forum text. These annotations therefore reflect the genuine semantic content of real user posts. Only in the second step the comments were synthesized: the LLM was instructed to lightly rephrase the original text and replace identifying entities with placeholders, while explicitly preserving the semantic content that the previously extracted annotations describe. The annotations serve as ground truth derived from real user content, and the synthesized text is a privacy-preserving reformulation of that same content.

To extract insights for each branch, LLM was applied to the data. Specifically, the OpenAI GPT-4.1 model was used in this study. This model was selected as it represented the most capable and up-to-date release in the GPT-4 family at the time of dataset construction, offering state-of-the-art performance on long-context understanding and reliable structured JSON output generation. Its competitive cost-per-token ratio relative to other frontier models also made large-scale annotation of 7130 branches practically feasible within the project's resource constraints. A query consists of a system prompt, an assistant prompt, and a user prompt.

The assistant prompt includes a description of the task, as well as the structure of the report that the model should return for each category, for example: 
\begin{quote}
\textit{"Please analyze the provided forum branch and generate a report based on all comments containing the following sections:
(1) UX Insight Summary: summarize the main content and implications on the user experience with the product, focusing on specific UX insights and main issues users have about the product; (2) etc."}
\end{quote}

In addition, the format of the output file was defined, for example: 

\textit{"Provide the report in this JSON structure:}
\begin{quote}
\begin{verbatim}
{ "insight_summary": " ", "user_expectations": " ", "gain_keywords": [ ], 
"pain_keywords": [ ], "feature_keywords": [ ], "overall_branch_sentiment": " ",
"os_system": { }, "severity_expectation_level": " " }"
\end{verbatim}
\end{quote}

The assistant prompt defines the role of the model as well as prior information about the structure of specific products, e.g., 

\begin{quote}
\textit{"You are a UX expert specializing in analyzing user feedback on forums about different products. You have received a branch of comments from a forum containing a user's question and answers from other users. Your task is to analyze the feedback and provide a report as per the provided structure. Be concise and provide only relevant information."}
\end{quote}

The user prompt displays all comments from a single branch, along with their metadata (time, views, rating), as well as user metadata (total number of posts, rating, rank).

\begin{figure}[H]
    \centering
    \includegraphics[width=0.6\columnwidth]{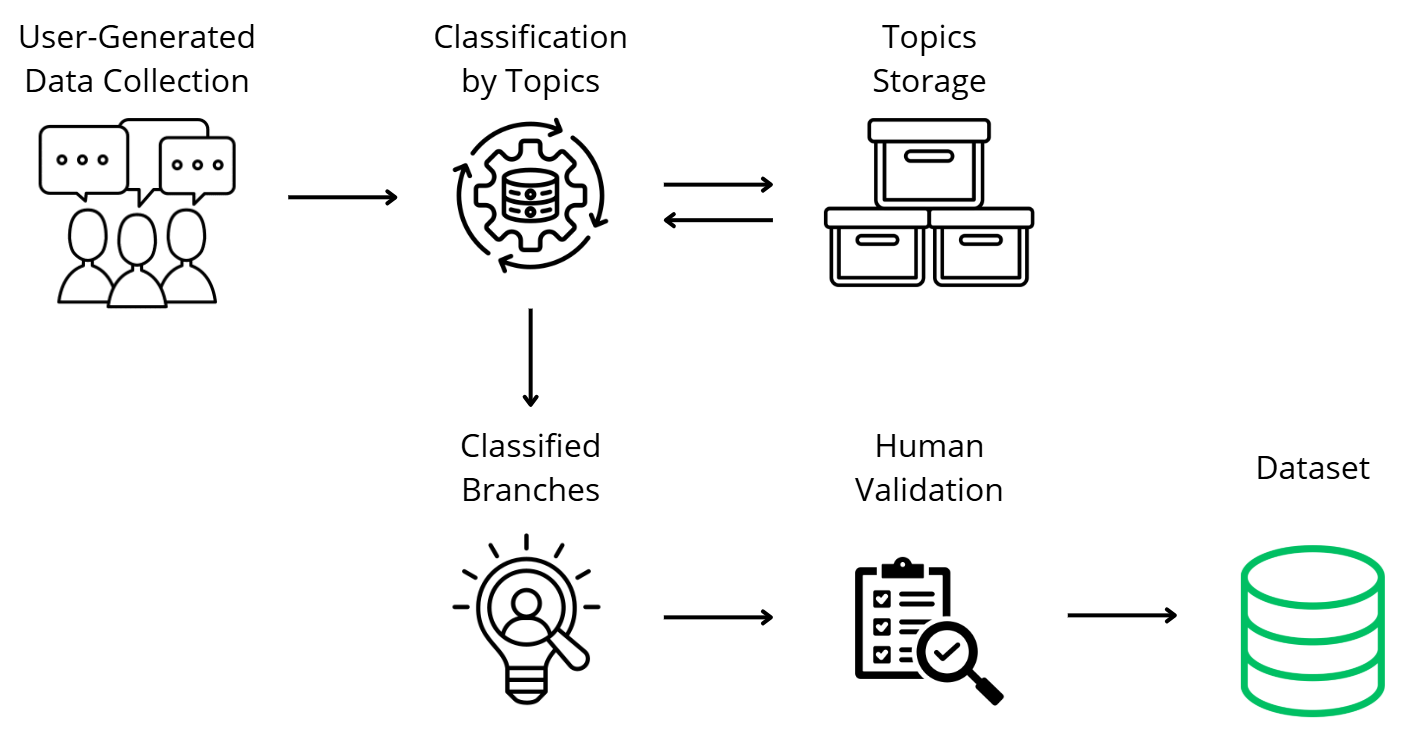}
    \caption{Overview of topic classification process}
    \label{fig:topics}
\end{figure}

Additionally, an iterative topic assignment process was applied to each thread branch to classify branches according to the types of problems experienced by, Fig.~\ref{fig:topics}. We independently extracted the topics from each branch with LLM. After each branch was processed, we saved the new topics in the storage. On every new branch, we first tried to select some of those. In case there was no match, we extracted the new one.

\subsection*{Anonymization}

The LLM was instructed to lightly rephrase each comment while simultaneously replacing all identifying information with typed placeholders. Specifically, the model was asked to substitute: company names with \texttt{[company\_name]}, product names with \texttt{[product\_name]}, article numbers with \texttt{[article\_no]}, version numbers with \texttt{[version\_no]}, personal names and user names with \texttt{[user\_name]} (including names appearing after sign-off phrases such as ``best regards'' or ``thank you''), URLs with \texttt{[url]}, and document names with \texttt{[document]}. The model was explicitly instructed that privacy is the primary concern and that all personal data must be anonymized, even in ambiguous cases.

Following the LLM pass, a classical pattern-matching step was applied to catch any residual identifiers the model may have missed. Regular expressions were used to detect and redact common personal data patterns, including email addresses, IP addresses, phone numbers, and remaining URL patterns not captured by the LLM.

A manual review was carried out on a sample of processed branches to verify that no identifying information remained and that the synthesised text preserved the original meaning and technical content.

\subsection*{Human Validation}

To assess the accuracy of the LLM-generated annotations, a structured manual validation was conducted on 32 purposively selected forum branches as part of a real product deployment evaluation, performed on the original unmodified posts before anonymization. Branches were chosen to cover six criteria: use cases, product features, competitors, operating systems, thread time span, and number of posts.

For each branch, three domain experts independently assessed 20 annotation fields \emph{before} consulting the LLM results to avoid foreknowledge bias, covering the full range of dataset fields, including use case, UX metrics, severity, sentiment, pain/feature keywords, branch type and status, and two experimental fields. Validation was conducted in three rounds: after each round, prompts were revised to address identified discrepancies, and the same sample was re-evaluated.

Agreement was classified using a three-level categorical scheme following established observer-agreement frameworks~\cite{LandisKoch1977,Cohen1968}: \emph{green} (exact agreement), \emph{yellow} (minor disagreement, e.g.\ some aspects not fully covered), and \emph{red} (clear disagreement). This scheme treats disagreements as not equally severe, consistent with the weighted-disagreement logic of Cohen~\cite{Cohen1968} and with structured
content-validity assessment frameworks~\cite{PolitBeck2006,PolitBeckOwen2007}.

Across all 640 outputs ($32 \times 20$ fields), 73\,\% were green, 14\,\%
yellow, and 13\,\% red, yielding an 87\,\% adequate-agreement rate.
Disagreements were not uniformly distributed: red ratings were concentrated in fields requiring contextual inference (branch status, experimental fields), while well-defined extraction fields (severity, sentiment) showed high agreement. Prompt revisions between rounds reduced systematic discrepancies, confirming that a substantial share of initial errors were addressable through prompt refinement.

\section*{Data records}

\begin{figure}[h!]
    \centering
    \includegraphics[width=0.9\columnwidth]{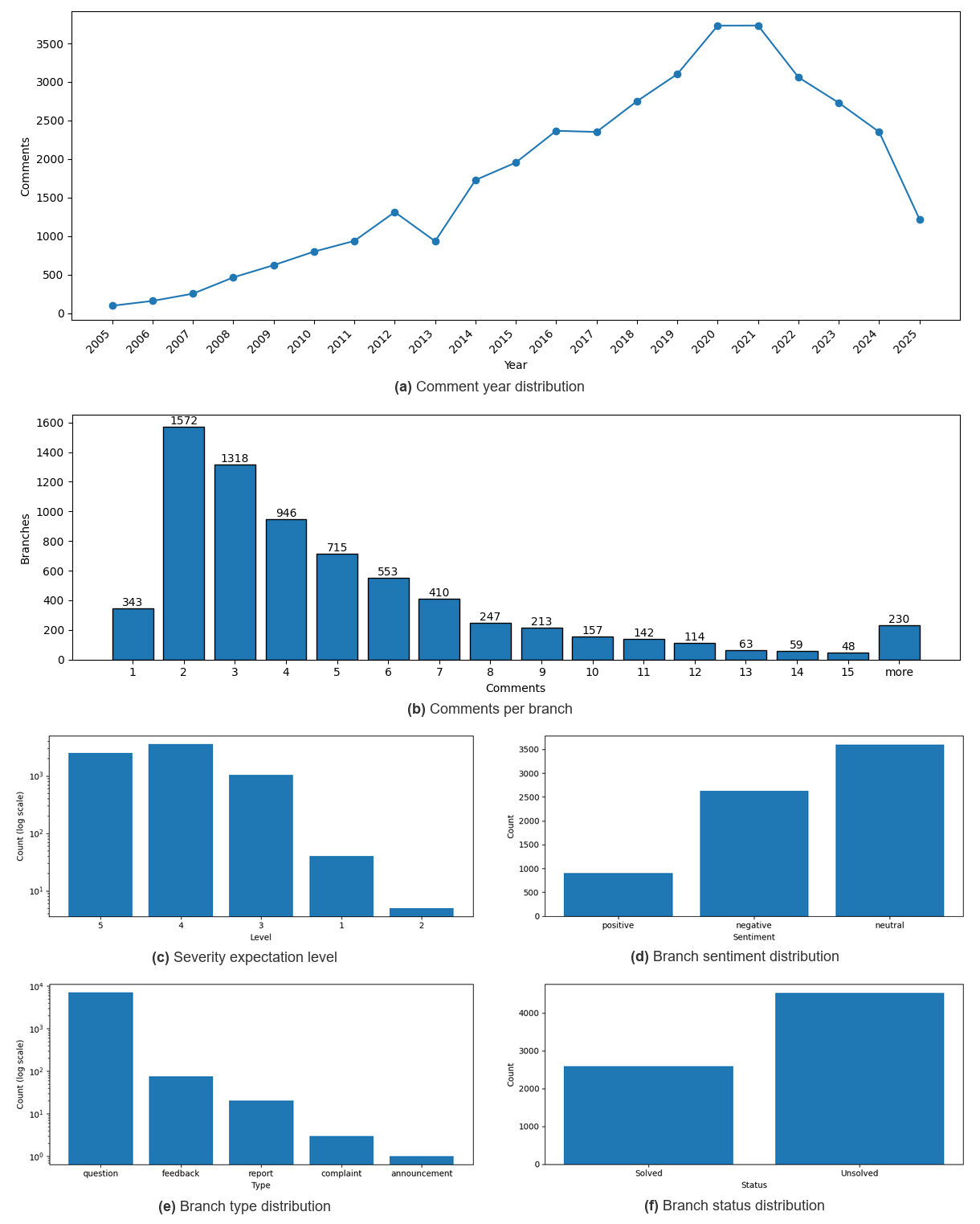}
    \caption{Distribution of the branches by year of the publication (a), number of comments (b), severity level (c), sentiment (d), branch type(e), and branch status (f).}
    \label{fig:distribution}
\end{figure}

The UXPID dataset is available at Zenodo \cite{Kulyabin2025UXPID}. The dataset consists of 7130 JSON files with more than 36k comments. Each of the JSON files represents a single forum branch consisting of at least one post and the related responses, summing up to 29k replies in total. The mean number of comments in the collected data is 5.1 with minimal of 1 and maximal of 54. The distributions of branches by year, number of comments, severity, sentiment, type, and branch status are shown in the Fig.~\ref{fig:distribution}. Dataset includes 4532 unsolved and 2598 solved branches (\texttt{branch\_status}). The \texttt{branch\_type} field records the structural nature of the initiating post: 7030 out of 7130 branches were initiated by a user question, which is typical of technical support forums where users seek solutions to problems. The remaining 100 branches correspond to other post types such as discussions or announcements.

Each JSON file consists of four parts: metadata, which provides general information about the branch; content, which contains processed and anonymized user comments; analysis, which presents obtained insights about the branch; and topics, which assign topic classes.

\begin{figure}[h]
    \centering
    \includegraphics[width=1\columnwidth]{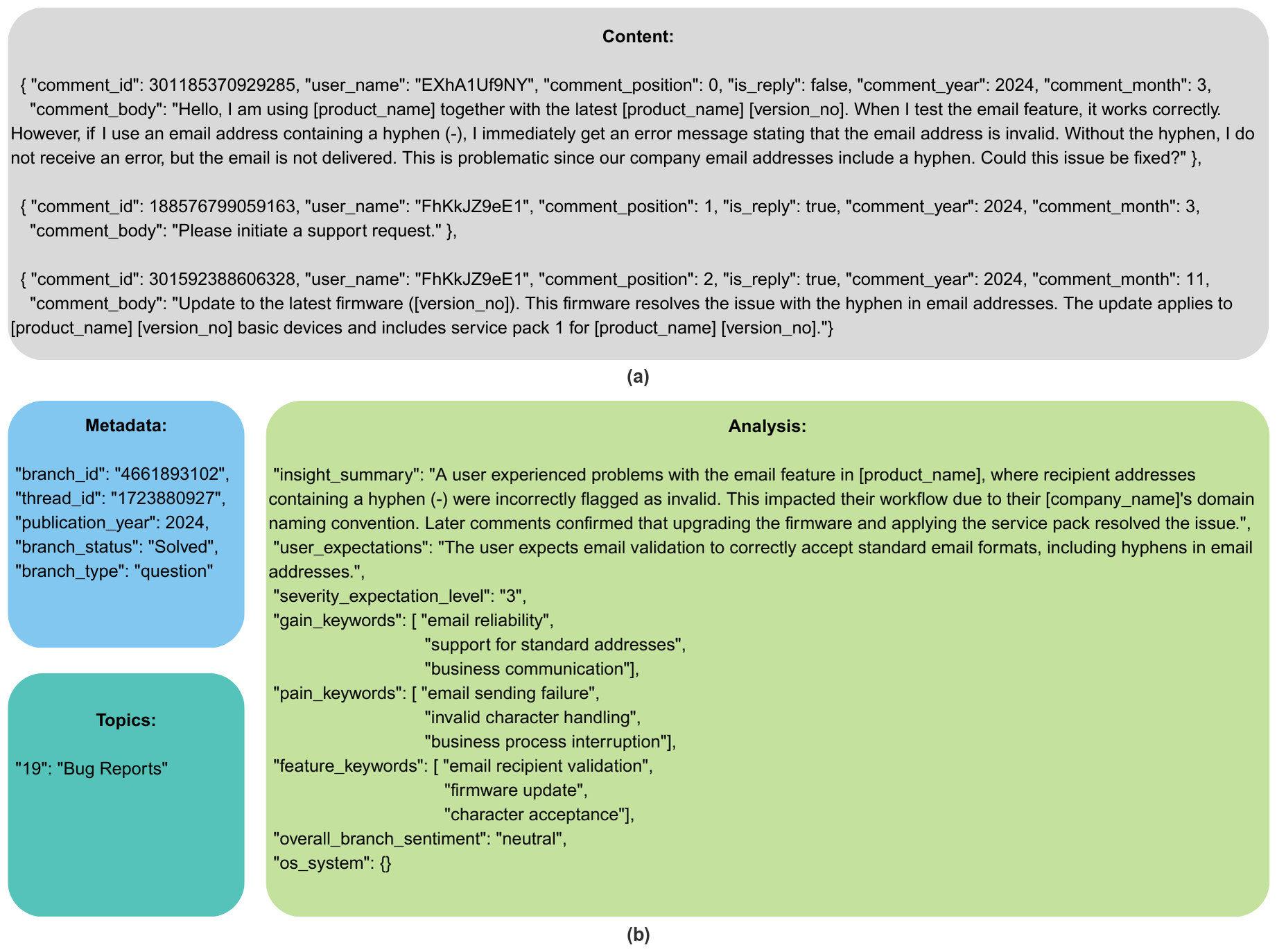}
    \caption{Example of the record structure from branch id 4661893102: content (a), metadata and analysis (b).}
    \label{fig:record_example}
\end{figure}

Metadata consists of the following fields: "branch\_id", "thread\_id", "publication\_year", "branch\_status", and "branch\_type". Content has "comment\_id", "user\_name" - anonymized but corresponds to the same users in the entire dataset, "comment\_position", "is\_reply" - which shows if the comment was an initial question or the answer, "comment\_year", "comment\_month", and "comment\_body" with the comment content. Analysis consists of "insight\_summary", "user\_expectations", "severity\_expectation\_level", "gain\_keywords", "pain\_keywords", "feature\_keywords", "overall\_branch\_sentiment", and "os\_system", where key is the system name (i.e., "Windows") and the value is an array with the mentioned versions (i.e., ["Windows 7", "Windows 10"]). Fig.~\ref{fig:record_example} presents a data record example.

\section*{Technical Validation}

To validate the dataset, we conducted a series of classification experiments using both a classical TF-IDF + Logistic Regression (TF-IDF+LR) baseline \cite{yun2005improved} and the DistilBERT \cite{sanh2020distilbertdistilledversionbert} language model across various scenarios, providing reference points for future work with potentially more advanced models. DistilBERT is a streamlined version of Bidirectional Encoder Representations from Transformers (BERT) \cite{devlin2018bert} that maintains impressive performance while requiring significantly less computational resources. This bidirectional transformer model was pre-trained on unlabeled text with two key objectives: predicting masked tokens within sentences and determining sentence relationships. The model's ability to process text from both directions simultaneously provides it with a comprehensive understanding of linguistic context.

DistilBERT achieves its efficiency through knowledge distillation, creating a more compact model that delivers faster inference times. During its pre-training phase, the model optimizes a triple loss objective encompassing language modeling loss, distillation loss, and cosine-distance loss. This approach allows DistilBERT to closely match the performance of larger transformer models while being more accessible. We sourced our implementation from Hugging Face's model repository.

As a classical baseline we employed a TF-IDF vectoriser paired with a one-vs-rest Logistic Regression classifier with balanced class weights. For multi-label topic classification, per-class decision thresholds were calibrated on a 15\% held-out slice of the training set to maximise per-class F1; for single-label sentiment classification a fixed 0.5 threshold was used. This pipeline requires no GPU and trains in seconds, providing a useful lower bound against which the transformer-based model can be compared.

Our validation focused on classification tasks related to branch metadata extraction. We explored four distinct scenarios by varying both input text and target variables: "Raw comments $\rightarrow$ Topic class"; "Raw comments $\rightarrow$ Overall sentiment; "Insights summary $\rightarrow$ Topic class"; "Insights summary $\rightarrow$ Overall sentiment".

The dataset was partitioned into a fixed train/test split using a stratification-free random split with a fixed seed equal to 42, yielding 5704 training branches (80\%) and 1426 test branches (20\%), with the split identifiers provided in the dataset repository as text files. An additional 10\% of the training set was held out as a validation subset during model selection and early stopping.

For all experiments, we utilized the distilbert-base-uncased configuration with a maximum sequence length of 512 tokens. The model architecture included 6 hidden layers with 768 dimensions each and a dropout rate of 0.4 to prevent overfitting. Our training and inference were performed on a Tesla T4 GPU with 16GB of memory.

The training process employed a batch size of 16 and a learning rate of 2e-5. While we allocated 100 epochs for each experiment, we implemented early stopping criteria to optimize training efficiency. To address class imbalance in our dataset, we applied class weighting techniques for both models. Complete model training parameters are available in the configuration file within our repository. Table~\ref{tab:test_topics} presents the inference results on the test subset across topic multi-label classification scenarios for both DistilBERT and TF-IDF+LR.

\begin{table}[h!]
\centering
\caption{Inference results across topic multi-label classification test scenarios. Support denotes the total number of ground-truth label assignments in the test set. ``All'' includes all 32 topic classes; ``$>50$\,Supports'' restricts evaluation to topic classes with more than 50 test-set instances, reducing noise from rare classes.}
\begin{tabular}{|c|c|c|c|c|c|c|c|c|c|}
\hline
\textbf{Task} & \textbf{Classes} & \textbf{Model} & \textbf{Support} & \textbf{AVG Pred.} & \textbf{AVG True} & \textbf{Precision} & \textbf{Recall} & \textbf{F1} & \textbf{AUC} \\
\hline
Summary $\rightarrow$ Topic & All & DistilBERT & 2410 & 1.6200 & 1.6900 & 0.7378 & 0.6667 & 0.6923 & 0.9797 \\
\hline
Summary $\rightarrow$ Topic & All & TF-IDF+LR & 2410 & 1.9299 & 1.6900 & 0.4625 & 0.5328 & 0.4799 & 0.9409 \\
\hline
Summary $\rightarrow$ Topic & $>50$\,Supp. & DistilBERT & 2012 & 1.6200 & 1.6900 & 0.7390 & 0.7009 & 0.7174 & 0.9698 \\
\hline
Summary $\rightarrow$ Topic & $>50$\,Supp. & TF-IDF+LR & 2012 & 1.9299 & 1.6900 & 0.5832 & 0.6505 & 0.6060 & 0.9099 \\
\hline
Comments $\rightarrow$ Topic & All & DistilBERT & 2410 & 1.5300 & 1.6900 & 0.6766 & 0.5629 & 0.6134 & 0.9405 \\
\hline
Comments $\rightarrow$ Topic & All & TF-IDF+LR & 2410 & 2.4635 & 1.6900 & 0.3391 & 0.5036 & 0.3765 & 0.8896 \\
\hline
Comments $\rightarrow$ Topic & $>50$\,Supp. & DistilBERT & 2012 & 1.5300 & 1.6900 & 0.6755 & 0.5362 & 0.5918 & 0.9633 \\
\hline
Comments $\rightarrow$ Topic & $>50$\,Supp. & TF-IDF+LR & 2012 & 2.4635 & 1.6900 & 0.4707 & 0.5772 & 0.5104 & 0.8526 \\
\hline
\end{tabular}
\label{tab:test_topics}
\end{table}

\begin{table}[h!]
\centering
\caption{Inference results across overall sentiment branch classification test scenarios. Support denotes the number of test-set branches with a valid sentiment label.}
\begin{tabular}{|c|c|c|c|c|c|c|c|}
\hline
\textbf{Task} & \textbf{Classes} & \textbf{Model} & \textbf{Support} & \textbf{Precision} & \textbf{Recall} & \textbf{F1} & \textbf{AUC} \\
\hline
Summary $\rightarrow$ Overall Sentiment & All & DistilBERT & 1425 & 0.8037 & 0.7866 & 0.7825 & 0.9331 \\
\hline
Summary $\rightarrow$ Overall Sentiment & All & TF-IDF+LR & 1425 & 0.6548 & 0.6598 & 0.6568 & 0.8423 \\
\hline
Summary $\rightarrow$ Overall Sentiment & All & DistilBERT & 1425 & 0.7945 & 0.7920 & 0.7923 & 0.9274 \\
\hline
Summary $\rightarrow$ Overall Sentiment & All & TF-IDF+LR & 1425 & 0.6280 & 0.6416 & 0.6340 & 0.8162 \\
\hline
\end{tabular}
\label{tab:test_sentiment}
\end{table}

The average number of true topic classes per branch is 1.69. DistilBERT predicts 1.62 and 1.53 labels per sample for the summary and raw-comment inputs, respectively, closely matching the ground truth, whereas TF-IDF+LR tends to over-predict. DistilBERT consistently outperforms the classical baseline, achieving an F1 of 0.6923 versus 0.4799 on summary-to-topic (All classes) and 0.6134 versus 0.3765 on comments-to-topic. When evaluation is restricted to the 10 topic classes with more than 50 test-set instances, TF-IDF+LR reaches an F1 of 0.6060 on summaries, indicating that much of the gap stems from rare classes where limited training signal hampers the bag-of-words model. AUC scores are closer, suggesting that the ranking quality of the classical baseline is reasonable, but its hard-label thresholding is less effective. The results of all metrics are predictably higher in the prediction scenario from Insights summaries, since topics were extracted from them when preparing the dataset.

Table~\ref{tab:test_sentiment} presents the inference results on the test subset across overall sentiment branch classification scenarios. DistilBERT achieves F1 scores of 0.7825 (summary) and 0.7923 (comments), while TF-IDF+LR reaches 0.6568 and 0.6340. Because user insights and sentiment were extracted from raw comments, the gap between the two input modalities is small for both models. Unlike topic classification, sentiment is a three-class single-label problem (neutral, negative, positive), which benefits bag-of-words features less than contextual embeddings, yielding a larger relative gap between the two models. The metrics for inference are generally higher than for the multi-label topic classification problem, as expected for a simpler label space.

The results show that the UXPID dataset advances AI and data-driven methods for technical forums and industrial product support. It contains realistic, well-annotated conversation branches that capture authentic user experiences in automation hardware and software. UXPID fills gaps in existing resources by adding topic clustering, severity annotations, and scenario-based insights. Its privacy-preserving design enables broad academic and commercial use. The dataset supports the development, training, and benchmarking of NLP models for sentiment analysis, issue detection, requirements extraction, and topic classification in technical user communities.

\subsection*{Fidelity Analysis}

\begin{figure}[h]
    \centering
    \includegraphics[width=0.9\columnwidth]{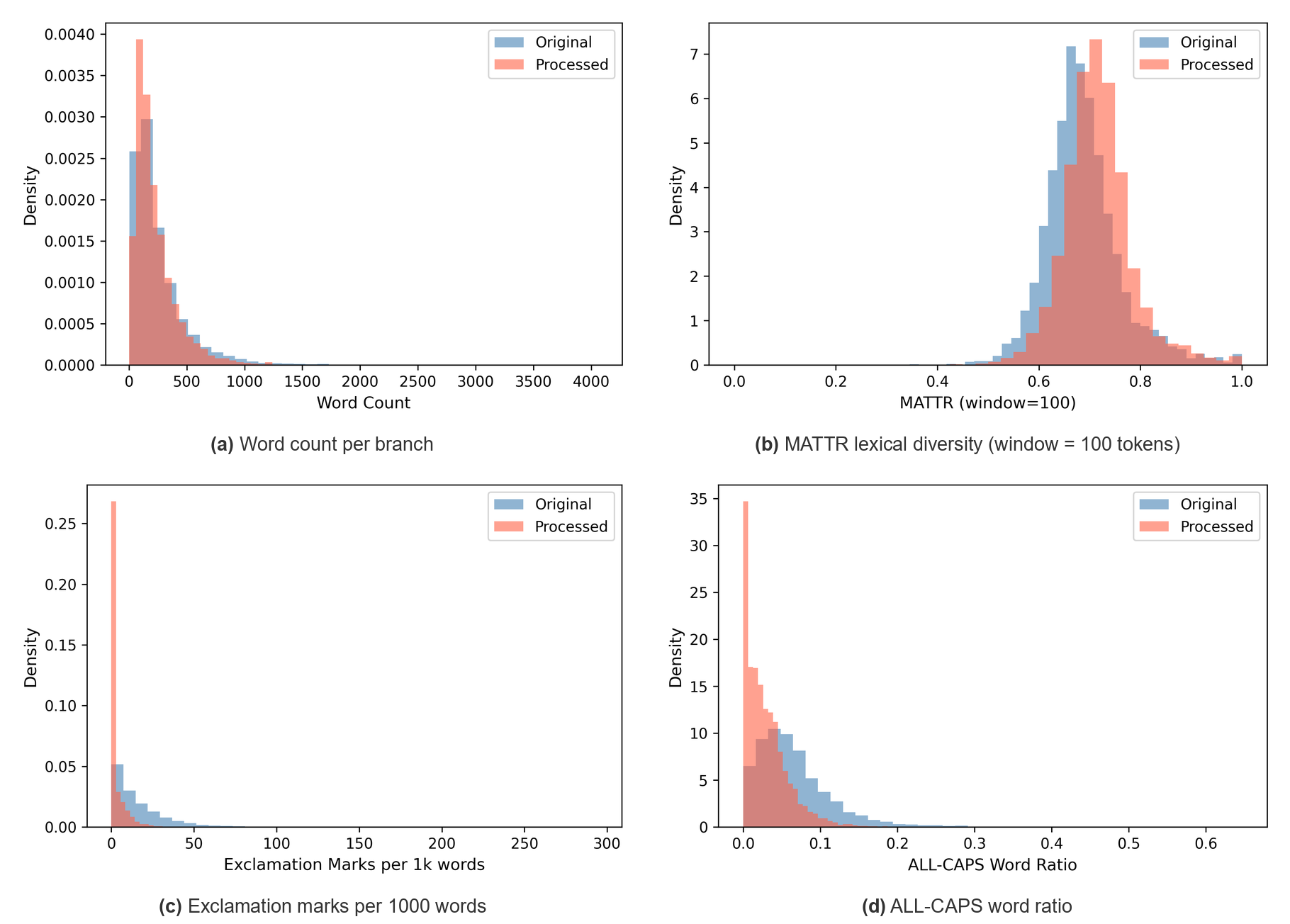}
    \caption{Density distributions of four linguistic features comparing original pre-synthesis (blue) and published post-synthesis (red) forum text branch pairs. Word count and MATTR distributions remain substantially overlapping, while exclamation mark frequency and ALL-CAPS ratio shift markedly toward zero after synthesis, reflecting the intentional formalisation of the surface register.}
    \label{fig:fidelity}
\end{figure}

To quantify the linguistic impact of the anonymization and synthesis step, we performed a paired statistical comparison between the original pre-synthesis forum text and the processed text across all branch pairs using the Wilcoxon signed-rank test throughout. Conversational structure was completely preserved: comment count, participant count, and reply ratio were identical across all pairs. Once punctuation counts were normalised per 1000 words to control for the overall 12.1\% reduction in word count (Fig.~\ref{fig:fidelity}a), question mark frequency changed by only $-4.5\%$ ($p = 0.027$), indicating that interrogative discourse structure is largely retained. Length-invariant lexical diversity (MATTR, Fig.~\ref{fig:fidelity}b) increased by a modest 4.3\% ($p < 0.0001$); because repeated anonymization placeholders such as \texttt{[product\_name]} suppress MATTR in the processed text, this figure is a conservative lower bound on the true vocabulary enrichment introduced by GPT-4.1.

Surface formality increased in a direction consistent with the anonymization design. Flesch Reading Ease decreased by 24.2\% (58.5 $\rightarrow$ 44.4), and average word length increased by 15.5\%, reflecting denser vocabulary. Informal markers were substantially reduced: exclamation marks fell by 84.7\% (15.9 $\rightarrow$ 2.4 per 1000 words, Fig.~\ref{fig:fidelity}c) and ALL-CAPS word ratio dropped by 55.9\% (Fig.~\ref{fig:fidelity}d), with a large fraction of processed branches reaching exactly zero ALL-CAPS words. All differences were statistically significant at $p < 0.0001$ unless otherwise stated. Because all annotations were extracted from the original text prior to synthesis, the ground-truth labels reflect authentic user intent. Users should note that models trained on UXPID may underlearn the role of emphatic punctuation and informal register in sentiment expression when generalising to unprocessed forum data.

The multi-faceted annotations open several concrete research directions. The pain/gain/feature keyword fields support fine-grained requirements elicitation beyond binary sentiment classification. The preserved conversational structure - reply chains, comment positions, and participant counts - enables research on problem-solving dynamics, such as which interaction patterns correlate with resolved issues. The dataset can also serve as a benchmark for comparing LLM-generated insights with other feedback channels, such as app store reviews, to study how technical forum language differs from consumer-facing expressions.

A particularly relevant application concerns industrial scalability. Large organisations accumulate millions of forum comments that remain largely unanalysed. LLM-based pipelines are currently the strongest option for extracting structured insights at this scale, but reprocessing all historical branches whenever new comments arrive incurs significant and recurring API costs. UXPID provides a resource for developing more efficient strategies - such as distilling smaller task-specific models from the LLM-generated labels, or using active learning to prioritise which branches need reprocessing. Key open questions include how closely lighter models trained on UXPID can replicate the original LLM annotations, and whether models transfer to other technical domains such as consumer electronics or enterprise software without domain-specific fine-tuning.

\section*{Data availability}

The dataset available at \href{https://zenodo.org/records/19055619}{https://zenodo.org/records/19055619}.

\section*{Code availability}

The custom code and models used to generate and process the dataset described in this paper are available at our GitHub repository:
\href{https://github.com/MikhailKulyabin/UXPID}{https://github.com/MikhailKulyabin/UXPID}.

\bibliography{references}

\begin{thebibliography}{10}
\urlstyle{rm}
\expandafter\ifx\csname url\endcsname\relax
  \def\url#1{\texttt{#1}}\fi
\expandafter\ifx\csname urlprefix\endcsname\relax\def\urlprefix{URL }\fi
\expandafter\ifx\csname doiprefix\endcsname\relax\def\doiprefix{DOI: }\fi
\providecommand{\bibinfo}[2]{#2}
\providecommand{\eprint}[2][]{\url{#2}}
\providecommand{\JournalTitle}[1]{#1}

\bibitem{maalej2025automated}
\bibinfo{author}{Maalej, W.}, \bibinfo{author}{Biryuk, V.}, \bibinfo{author}{Wei, J.} \& \bibinfo{author}{Panse, F.}
\newblock \bibinfo{title}{On the automated processing of user feedback}.
\newblock In \emph{\bibinfo{booktitle}{Handbook on Natural Language Processing for Requirements Engineering}}, \bibinfo{pages}{279--308} (\bibinfo{publisher}{Springer}, \bibinfo{year}{2025}).

\bibitem{zhang2025less}
\bibinfo{author}{Zhang, J.} \emph{et~al.}
\newblock \bibinfo{journal}{\bibinfo{title}{Less is more: On the importance of data quality for unit test generation}}.
\newblock {\emph{\JournalTitle{Proceedings of the ACM on Software Engineering}}} \textbf{\bibinfo{volume}{2}}, \bibinfo{pages}{1293--1316},  (\bibinfo{year}{2025}).

\bibitem{ISO9241-210}
\bibinfo{title}{Ergonomics of human-system interaction — part 210: Human-centred design for interactive systems},  (\bibinfo{year}{2019}).

\bibitem{lopez_2023_re}
\bibinfo{author}{Tasnim, M.}, \bibinfo{author}{Rayhan, M.}, \bibinfo{author}{Zhang, Z.} \& \bibinfo{author}{Poranen, T.}
\newblock \bibinfo{title}{A systematic literature review on requirements engineering practices and challenges in open-source projects}.
\newblock In \emph{\bibinfo{booktitle}{2023 49th Euromicro Conference on Software Engineering and Advanced Applications (SEAA)}}, \bibinfo{pages}{278--285}, \url{10.1109/SEAA60479.2023.00050} (\bibinfo{year}{2023}).

\bibitem{sommerville1997requirements}
\bibinfo{author}{Sommerville, I.} \& \bibinfo{author}{Sawyer, P.}
\newblock \emph{\bibinfo{title}{Requirements engineering: a good practice guide}} (\bibinfo{publisher}{John Wiley \& Sons, Inc.}, \bibinfo{year}{1997}).

\bibitem{Harte2017}
\bibinfo{author}{Harte, R.} \emph{et~al.}
\newblock \bibinfo{journal}{\bibinfo{title}{A human-centered design methodology to enhance the usability, human factors, and user experience of connected health systems: A three-phase methodology}}.
\newblock {\emph{\JournalTitle{JMIR Human Factors}}} \textbf{\bibinfo{volume}{4}}, \bibinfo{pages}{e8}, \url{10.2196/humanfactors.5443},  (\bibinfo{year}{2017}).

\bibitem{Sauer02102020}
\bibinfo{author}{Sauer, J.}, \bibinfo{author}{Sonderegger, A.} \& \bibinfo{author}{Schmutz, S.}
\newblock \bibinfo{journal}{\bibinfo{title}{Usability, user experience and accessibility: towards an integrative model}}.
\newblock {\emph{\JournalTitle{Ergonomics}}} \textbf{\bibinfo{volume}{63}}, \bibinfo{pages}{1207--1220}, \url{10.1080/00140139.2020.1774080},  (\bibinfo{year}{2020}).
\newblock \bibinfo{note}{PMID: 32450782}, \eprint{https://doi.org/10.1080/00140139.2020.1774080}.

\bibitem{soares_user_2011}
\bibinfo{author}{Soares, M. D.~S.}, \bibinfo{author}{Vrancken, J.} \& \bibinfo{author}{Verbraeck, A.}
\newblock \bibinfo{journal}{\bibinfo{title}{User requirements modeling and analysis of software-intensive systems}} \textbf{\bibinfo{volume}{84}}, \bibinfo{pages}{328--339}, \url{10.1016/j.jss.2010.10.020},  (\bibinfo{year}{2011}).

\bibitem{kang2020natural}
\bibinfo{author}{Kang, Y.}, \bibinfo{author}{Cai, Z.}, \bibinfo{author}{Tan, C.-W.}, \bibinfo{author}{Huang, Q.} \& \bibinfo{author}{Liu, H.}
\newblock \bibinfo{journal}{\bibinfo{title}{Natural language processing (nlp) in management research: A literature review}}.
\newblock {\emph{\JournalTitle{Journal of Management Analytics}}} \textbf{\bibinfo{volume}{7}}, \bibinfo{pages}{139--172},  (\bibinfo{year}{2020}).

\bibitem{hirschberg2015advances}
\bibinfo{author}{Hirschberg, J.} \& \bibinfo{author}{Manning, C.~D.}
\newblock \bibinfo{journal}{\bibinfo{title}{Advances in natural language processing}}.
\newblock {\emph{\JournalTitle{Science}}} \textbf{\bibinfo{volume}{349}}, \bibinfo{pages}{261--266},  (\bibinfo{year}{2015}).

\bibitem{just2024natural}
\bibinfo{author}{Just, J.}
\newblock \bibinfo{journal}{\bibinfo{title}{Natural language processing for innovation search--reviewing an emerging non-human innovation intermediary}}.
\newblock {\emph{\JournalTitle{Technovation}}} \textbf{\bibinfo{volume}{129}}, \bibinfo{pages}{102883},  (\bibinfo{year}{2024}).

\bibitem{laurer2024less}
\bibinfo{author}{Laurer, M.}, \bibinfo{author}{Van~Atteveldt, W.}, \bibinfo{author}{Casas, A.} \& \bibinfo{author}{Welbers, K.}
\newblock \bibinfo{journal}{\bibinfo{title}{Less annotating, more classifying: Addressing the data scarcity issue of supervised machine learning with deep transfer learning and bert-nli}}.
\newblock {\emph{\JournalTitle{Political Analysis}}} \textbf{\bibinfo{volume}{32}}, \bibinfo{pages}{84--100},  (\bibinfo{year}{2024}).

\bibitem{osman_quality_2019}
\bibinfo{author}{Osman, A.}, \bibinfo{author}{Salim, N.} \& \bibinfo{author}{Saeed, F.}
\newblock \bibinfo{journal}{\bibinfo{title}{Quality dimensions features for identifying high-quality user replies in text forum threads using classification methods}} \textbf{\bibinfo{volume}{14}}, \bibinfo{pages}{e0215516}, \url{10.1371/journal.pone.0215516}.

\bibitem{castelli2019techqa}
\bibinfo{author}{Castelli, V.} \emph{et~al.}
\newblock \bibinfo{journal}{\bibinfo{title}{The techqa dataset}}.
\newblock {\emph{\JournalTitle{arXiv preprint arXiv:1911.02984}}}  (\bibinfo{year}{2019}).

\bibitem{sonali_fr_nfr_dataset_2024}
\bibinfo{author}{Sonali, S.}
\newblock \bibinfo{title}{{FR}\_nfr\_dataset}, \url{10.17632/4YSX9FYZV4.1},  (\bibinfo{year}{2024}).

\bibitem{ferrari2017pure}
\bibinfo{author}{Ferrari, A.}, \bibinfo{author}{Spagnolo, G.~O.} \& \bibinfo{author}{Gnesi, S.}
\newblock \bibinfo{title}{Pure: A dataset of public requirements documents}.
\newblock In \emph{\bibinfo{booktitle}{2017 IEEE 25th international requirements engineering conference (RE)}}, \bibinfo{pages}{502--505} (\bibinfo{organization}{IEEE}, \bibinfo{year}{2017}).

\bibitem{Bozyigit_2023_r9j6-nd62-23}
\bibinfo{author}{Bozyigit, F.} \emph{et~al.}
\newblock \bibinfo{title}{Dataset for: Text requirements to models}, \url{10.21227/r9j6-nd62},  (\bibinfo{year}{2023}).

\bibitem{mekala2021classifying}
\bibinfo{author}{Mekala, R.~R.}, \bibinfo{author}{Irfan, A.}, \bibinfo{author}{Groen, E.~C.}, \bibinfo{author}{Porter, A.} \& \bibinfo{author}{Lindvall, M.}
\newblock \bibinfo{title}{Classifying user requirements from online feedback in small dataset environments using deep learning}.
\newblock In \emph{\bibinfo{booktitle}{2021 IEEE 29th International requirements engineering conference (RE)}}, \bibinfo{pages}{139--149} (\bibinfo{organization}{IEEE}, \bibinfo{year}{2021}).

\bibitem{kadebu2023classification}
\bibinfo{author}{Kadebu, P.}, \bibinfo{author}{Sikka, S.}, \bibinfo{author}{Tyagi, R.~K.} \& \bibinfo{author}{Chiurunge, P.}
\newblock \bibinfo{journal}{\bibinfo{title}{A classification approach for software requirements towards maintainable security}}.
\newblock {\emph{\JournalTitle{Scientific African}}} \textbf{\bibinfo{volume}{19}}, \bibinfo{pages}{e01496},  (\bibinfo{year}{2023}).

\bibitem{neo_user_2024}
\bibinfo{author}{Neo, G.}, \bibinfo{author}{Moura, J.}, \bibinfo{author}{Almeida, H.}, \bibinfo{author}{Neo, A.} \& \bibinfo{author}{Freitas~Júnior, O.}
\newblock \bibinfo{title}{User story tutor ({UST}) to support agile software developers:}.
\newblock In \emph{\bibinfo{booktitle}{Proceedings of the 16th International Conference on Computer Supported Education}}, \bibinfo{pages}{51--62}, \url{10.5220/0012619200003693} (\bibinfo{publisher}{{SCITEPRESS} - Science and Technology Publications}, \bibinfo{year}{2024}).

\bibitem{Kulyabin2025UXPID}
\bibinfo{author}{Kulyabin, M.} \emph{et~al.}
\newblock \bibinfo{title}{User experience perception insights dataset (uxpid)}, \url{10.5281/zenodo.19055619},  (\bibinfo{year}{2025}).
\newblock \bibinfo{note}{[Data set]}.

\bibitem{LandisKoch1977}
\bibinfo{author}{Landis, J.~R.} \& \bibinfo{author}{Koch, G.~G.}
\newblock \bibinfo{journal}{\bibinfo{title}{The measurement of observer agreement for categorical data}}.
\newblock {\emph{\JournalTitle{Biometrics}}} \textbf{\bibinfo{volume}{33}}, \bibinfo{pages}{159--174}, \url{10.2307/2529310},  (\bibinfo{year}{1977}).

\bibitem{Cohen1968}
\bibinfo{author}{Cohen, J.}
\newblock \bibinfo{journal}{\bibinfo{title}{Weighted kappa: Nominal scale agreement with provision for scaled disagreement or partial credit}}.
\newblock {\emph{\JournalTitle{Psychological Bulletin}}} \textbf{\bibinfo{volume}{70}}, \bibinfo{pages}{213--220}, \url{10.1037/h0026256},  (\bibinfo{year}{1968}).

\bibitem{PolitBeck2006}
\bibinfo{author}{Polit, D.~F.} \& \bibinfo{author}{Beck, C.~T.}
\newblock \bibinfo{journal}{\bibinfo{title}{The content validity index: Are you sure you know what's being reported? critique and recommendations}}.
\newblock {\emph{\JournalTitle{Research in Nursing \& Health}}} \textbf{\bibinfo{volume}{29}}, \bibinfo{pages}{489--497}, \url{10.1002/nur.20147},  (\bibinfo{year}{2006}).

\bibitem{PolitBeckOwen2007}
\bibinfo{author}{Polit, D.~F.}, \bibinfo{author}{Beck, C.~T.} \& \bibinfo{author}{Owen, S.~V.}
\newblock \bibinfo{journal}{\bibinfo{title}{Is the {CVI} an acceptable indicator of content validity? appraisal and recommendations}}.
\newblock {\emph{\JournalTitle{Research in Nursing \& Health}}} \textbf{\bibinfo{volume}{30}}, \bibinfo{pages}{459--467}, \url{10.1002/nur.20199},  (\bibinfo{year}{2007}).

\bibitem{yun2005improved}
\bibinfo{author}{Yun-tao, Z.}, \bibinfo{author}{Ling, G.} \& \bibinfo{author}{Yong-cheng, W.}
\newblock \bibinfo{journal}{\bibinfo{title}{An improved tf-idf approach for text classification}}.
\newblock {\emph{\JournalTitle{Journal of Zhejiang University-Science A}}} \textbf{\bibinfo{volume}{6}}, \bibinfo{pages}{49--55},  (\bibinfo{year}{2005}).

\bibitem{sanh2020distilbertdistilledversionbert}
\bibinfo{author}{Sanh, V.}, \bibinfo{author}{Debut, L.}, \bibinfo{author}{Chaumond, J.} \& \bibinfo{author}{Wolf, T.}
\newblock \bibinfo{title}{Distilbert, a distilled version of bert: smaller, faster, cheaper and lighter},  (\bibinfo{year}{2020}).
\newblock \eprint{1910.01108}.

\bibitem{devlin2018bert}
\bibinfo{author}{Devlin, J.}, \bibinfo{author}{Chang, M.-W.}, \bibinfo{author}{Lee, K.} \& \bibinfo{author}{Toutanova, K.}
\newblock \bibinfo{title}{Bert: Pre-training of deep bidirectional transformers for language understanding},  (\bibinfo{year}{2018}).
\newblock \eprint{1810.04805}.

\end{thebibliography}

\section*{Author contributions statement}

Data collection, M.K. and J.J.; conceptualization, J.J., M.K. and N.N.M.P; software, M.K. and J.J.; writing-original draft preparation, M.K., J.J., and N.N.M.P.; writing---review and editing, C.U., F.R. and F.P.; supervision, F.P.; H.O., and J.B. All authors have read and agreed to the published version of the manuscript.

\section*{Competing interests}

The authors declare that they have no known competing financial interests or personal relationships that could have appeared to influence the work reported in this paper.

\end{document}